\pdfoutput=1

\documentclass[11pt]{article}

\usepackage{acl}
\usepackage{graphicx}
\usepackage{times}
\usepackage{latexsym}
\usepackage[frozencache,cachedir=minted-cache]{minted}
\usepackage[T1]{fontenc}

\usepackage[utf8]{inputenc}

\usepackage{microtype}
\usepackage[compact]{titlesec}

\usepackage{soul}
\usepackage{xspace}

\expandafter\def\expandafter\UrlBreaks\expandafter{\UrlBreaks
  \do\a\do\b\do\c\do\d\do\e\do\f\do\g\do\h\do\i\do\j%
  \do\k\do\l\do\m\do\n\do\o\do\p\do\q\do\r\do\s\do\t%
  \do\u\do\v\do\w\do\x\do\y\do\z\do\A\do\B\do\C\do\D%
  \do\E\do\F\do\G\do\H\do\I\do\J\do\K\do\L\do\M\do\N%
  \do\O\do\P\do\Q\do\R\do\S\do\T\do\U\do\V\do\W\do\X%
  \do\Y\do\Z}

\title{Twitter-Demographer \\  A Flow-based Tool to Enrich Twitter Data}

\author{Federico Bianchi\\
  Bocconi University\\
  Milano, Italy \\
  \texttt{f.bianchi@unibocconi.it} \\\And
  Vincenzo Cutrona\\
  SUPSI \\
  Lugano, Switzerland \\
  \texttt{vincenzo.cutrona@supsi.ch} \\\AND
  Dirk Hovy\\
  Bocconi University\\
  Milano, Italy \\
  \texttt{dirk.hovy@unibocconi.it} 
  }

\newcommand{\tool}{Twitter-Demographer\xspace}

\begin{document}
\maketitle

\begin{abstract}
Twitter data have become essential to Natural
Language Processing (NLP) and social science research, driving various scientific discoveries in recent years. 
However, the textual data alone are often not enough to conduct studies: especially social scientists need more variables to perform their analysis and control for various factors. How we augment this information, such as users' location, age, or tweet sentiment, has ramifications for anonymity and reproducibility, and requires dedicated effort. This paper describes Twitter-Demographer, a simple, flow-based tool to enrich Twitter data with additional information about tweets and users. Twitter-Demographer is aimed at NLP practitioners and (computational) social scientists who want to enrich their datasets with aggregated information, facilitating reproducibility, and providing algorithmic privacy-by-design measures for pseudo-anonymity. We discuss our design choices, inspired by the flow-based programming paradigm, to use black-box components that can easily be chained together and extended. We also analyze the ethical issues related to the use of this tool, and the built-in measures to facilitate pseudo-anonymity.
\end{abstract}

\section{Introduction}
Twitter data are at the heart of NLP and social science research~\cite{steinert2018twitter}, used to study policy and decision making, and to better understand the consequences of public opinion.
Its accessibility and the variety and abundance of the data make Twitter one of the most fruitful sources to experiment with new NLP methods, and to generate insights into societal behavior \cite[e.g.,][]{munger2017tweetment}. Given that 199 million people communicate on Twitter daily,\footnote{\url{https://s22.q4cdn.com/826641620/files/doc_financials/2021/q1/Q1'21-Shareholder-Letter.pdf}} it becomes fundamental to find ways to better interpret this information. 

However, to control for the effects of various covariates, to stratify the data into sensible subgroups, and to assess their reliability, researchers often need more than the pure text data. Social sciences typically require a recourse to external variables like age or location to control for confounds. In addition, NLP research has shown that integrating socio-demographic information can improve a wide range of classification tasks \cite{volkova-etal-2013-exploring,hovy-2015-demographic,lynn2017human,li2018towards,hovy-yang-2021-importance}.
By default, this information is not available,
and a wide range of NLP tools have been developed to infer measures from the text \cite[i.e., sentiment, syntactic structure:][inter alia]{balahur-2013-sentiment,kong-etal-2014-dependency} and user \cite[age, gender, income, person or company][inter alia]{preotiuc-pietro-etal-2015-analysis,wang2019demographic}.

Here, we introduce \tool, a tool that provides a simple and extensible interface for NLP and social science researchers. Starting from tweet ids (the common way to share Twitter data), the tool hydrates the original text, and can enrich it with additional information like the sentiment of the tweets, topics, or estimated demographic information of the author, using existing tools.
\tool builds on previous research \cite[e.g.,][]{wang2019demographic,bianchi-etal-2021-pre,barbieri-etal-2020-tweeteval,wolf-etal-2020-transformers}, but puts all these efforts together in one simple tool that can be used with little effort. \tool can be applied to extract information from different languages, as its default components are either multi-lingual or language-independent.\footnote{Note, however, that the use of language-specific classifiers might restrict the usage to specific languages.} \tool has a simple API that can be used to quickly add user-defined components quickly and effectively. 

One of our goals is to provide and enforce the generation of \textbf{reproducible data enrichment pipelines} (i.e., they can be shared and produce the same results if components are kept the same). With data enrichment we mean the process of extending a dataset, e.g., adding new inferred properties, or disambiguating its content~\cite{cutrona2019semantically}.
Our flow-based infrastructure makes it easy to produce and share pipelines with other researchers to reconstruct the datasets.

Most importantly, inferring user-attributes, even for research purposes, poses a privacy issue. We implement several algorithmic \textbf{privacy-by-design} solutions to facilitate \textbf{pseudo-anonymity} of the users, and to reduce the chance that their personal data or identifiers can be used to identify natural persons.

We believe that \tool can help (computational) social scientists wanting to analyze properties of their datasets in more depth, and provide NLP practitioners with a unified way to enrich and share data.

\paragraph{Contributions}
We introduce a new tool, \tool, to enrich datasets of tweets with additional information. The extensible tool enables NLP practitioners and computational social scientists to quickly adapt their own datasets with the features required for a specific analysis. \tool encodes the resulting enrichment pipeline in a stable, shareable, and reproducible format and implements privacy-by-design principles.

\section{The flow-based paradigm}
The flow-based paradigm is helpful for data handling because it allows users to easily combine different black-box components in many different ways, fitting different requirements time by time. Each component implements a specific task, it takes some inputs and returns some outputs. Many solutions employ this kind of paradigm (e.g., Apache NiFi\footnote{\url{https://nifi.apache.org/}}). These solutions are directed at experts like data engineers because they require some knowledge about the low-level details (e.g., how to handle data sources, how to manage data streams, event-based executions).

However, the advantage of this paradigm is that users do not have to know the intrinsic logic of each block (hence black-box). They only have to focus on combining these blocks to ensure the proper mapping between inputs/outputs of consecutive blocks. Indeed, the main disadvantages of manually building these pipelines are that (i) they require massive effort to be defined; (ii) they are sensitive to various hurdles, e.g., what happens if we cannot find one tweet or its location is unavailable? (iii) they are error-prone, with minor errors possibly tearing down entire pipelines, e.g., what happens if a Web service changes its exchange data format, or is no longer available?

\tool has been imagined as a low coupled set of components that operates on a dataset in tabular format (e.g., a Pandas DataFrame). Each component takes the dataset as input, applies some operations on it (e.g., adding columns), and returns the modified dataset. Components can be integrated into pipelines: we aim for high cohesion and low coupling principles to reduce possible errors at the component level. Each component exposes a set of required inputs (i.e., columns that must be contained in the input dataset) and a set of generated outputs (i.e., names of the new columns added to the dataset). Using this information, we can chain different components together to introduce dependencies (e.g., to run the sentiment analysis classifier, we need first to query Twitter and create a new column containing the text of tweets). Exposing the input and the outputs allows for the consistency between different components to be checked beforehand to avoid compatibility issues.

The flow-based setup makes it possible to replace any component with another one implementing the same task with a different logic, as long as the new component respects the communication interface (i.e., expected inputs and generated outputs). It is worth noting that the paradigm does not force a specific absolute order between components: a component requiring some columns as input (e.g., $\langle a, b\rangle$) must be placed in any position after the components generating such columns.

The goal of \tool is two-fold: 1) providing an easy-to-use interface for data enrichment and 2) providing a system that allows users to re-use existing components that are already implemented in \tool easily.

\section{Twitter-Demographer}

\begin{figure*}[ht!]
  \centering  \includegraphics[width=1\textwidth]{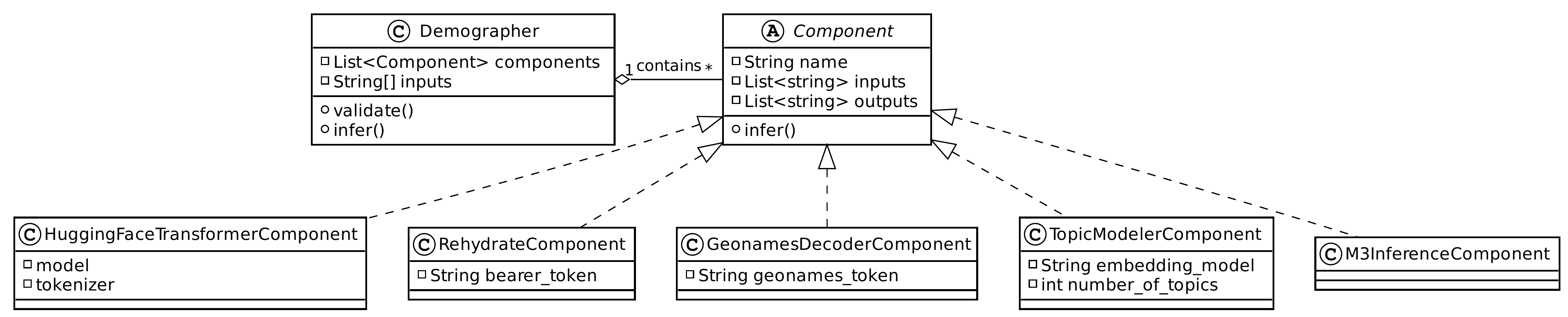}
  \caption{The UML class diagram of the current \tool setup. \textit{Demographer} is the main class that handles the execution of the different \textit{Component}s. \textit{Component} is an abstract class that defines required inputs and produced outputs, as well as an abstract \textit{infer()} methods that has to be implemented by its subclasses. Current available implementations of the Component class are reported in the UML diagram.}
  \label{fig:uml}
\end{figure*}

We show the class diagram of \tool in Figure~\ref{fig:uml}. While Listing~\ref{listing:scriptino} shows an example application of the tool. Line 2 instantiates the Demographer object, that is responsible for handling the entire pipeline (i.e., it also performs compatibility checks on components). Lines 4-6 show the instantiation of the different data augmentation components that will be used in the pipeline (a rehydration component to collect additional information from the tweets, a Geonames location decoder, and a sentiment classifier). Lines 9-11 add the components to the demographer object, creating the enrichment pipeline. Finally, line 14 runs the entire pipeline on the data, generating the enriched dataset.

\begin{listing}
\begin{minted}[fontsize=\footnotesize,numbers=left]{python}

# create the demographer object
demo = Demographer()

re = Rehydrate(token)
me = GeoNamesDecoder(user_name)
st = SentimentClassifier(model_name)

# add the components
demo.add_component(re)
demo.add_component(me)
demo.add_component(st)

# run the pipeline
new_data = demo.infer(data)
\end{minted}
\caption{Example of \tool basic usage. `data` variable is a simple DataFrame with one column containing the tweet ids.}
\label{listing:scriptino}
\end{listing}

We anyway guarantee the flexibility to allow new components to be implemented. A Component (Listing~\ref{listing:component_class}) is a simple abstract class that can be inherited and implemented easily: introducing a custom classification pipeline requires only to add a custom classifier to the pipeline, which inherits this class and implements the methods that handle inputs, outputs and the method to run the inference on the data.

\begin{listing}
\begin{minted}[fontsize=\footnotesize]
{python}
class Component(ABC):

    def __init__(self):
        self.outputs = self.outputs()

    @abc.abstractmethod
    def outputs(self):
        pass

    @abc.abstractmethod
    def inputs(self):
        pass

    @abc.abstractmethod
    def infer(self, *args):
        pass
        
\end{minted}
\caption{The implementation of the Component abstract class. `inputs`, `outputs` and `infer` are all abstract methods that have to be implemented by the inheriting classes.}
\label{listing:component_class}
\end{listing}

Inputs and outputs are exploited by Demographer to handle the control over the chain of possible components that can be added. A component cannot be added to a pipeline if it requires inputs that are not available in the original data, or that are not generated by previous components. For the sake of providing people with a simple system to extend, the current implementation of \tool represents these variables as lists of strings representing names of columns in data. As a next step, we will improve the current implementation by adopting a pure OOP point of view (i.e., inputs and outputs will turn into interfaces, with configurable parameters).

Listing~\ref{listing:user_defined_component} shows instead an example of implemented classifier; this is similar to how we have implemented some of our components, however, we report it also to show that this part of the pipeline can be used by interested researchers as an example of code to extend to support custom behaviors in \tool.

\begin{listing}
\begin{minted}[fontsize=\footnotesize]
{python}
class UserClassifier(Component):

    def __init__(self, model):
        super().__init__()
        self.m = model

    def outputs(self):
        return ["sentiment"]

    def inputs(self):
        return ["text"]

    def infer(self, data):
        return {"sentiment" : 
            self.m.predict(data["text"])}
        
\end{minted}
\caption{A user-defined component for sentiment classification of tweets. Users can add their own classifiers to the pipeline by wrapping them inside the Component abstraction.}
\label{listing:user_defined_component}
\end{listing}

\tool saves the intermediate computation steps, right after each component has been executed, to handle down-streaming unexpected errors (e.g., lost internet connection). In those situations, the computation can be restarted from checkpoints.

\subsection{Components}
\tool is a container of components, and can be extended as they are provided by the community. The current version of \tool is shipped with the following default components wrapped inside:

\begin{itemize}
    \item Basic ReHydration Component based on Twitter API v2. This components handles the retrieval of all the information that can be collected on Twitter from the single tweet id. It requires a Twitter API key.
    \item GeoNames Localizer.\footnote{\url{https://geocoder.readthedocs.io/}} A tool for geolocalizing users based on the location (e.g., address, state, and/or country) they manually write in the Twitter profile. This process is less precise than the geolocation given by Twitter, but also much more frequent: users often fill this field in their profile and thus it is a viable source of information. This localizer outputs the detected country and address.
    \item HuggingFace Transformer Classifier. A wrapper that can be used to use any classifier defined in the HuggingFace transformer library. With this wrapper, any classification module from HuggingFace can be used to classify the data (e.g., Hate Speech detection, Sentiment Analysis).
    \item Topic Modeling. A topic modeler based on Contextualized Topic Models~\cite{bianchi-etal-2021-cross,bianchi-etal-2021-pre} that also works on multi-lingual data. This topic modeling pipeline applies minor pre-processing by filtering infrequent words and removing links, users can select the number of topics they want to use to model the data.
    \item Gender and Age Predictor. A wrapper around the M3 classifier~\cite{wang2019demographic}\footnote{See Section \ref{sec:anonymity} and Section~\ref{sec:outlook} for a discussion of privacy by design and limitations} that can be used to predict binary gender, age group (i.e, >=40, 30-39, 19-29, <18) and identifies if the twitter account is an organization profile or not.
\end{itemize}

Some components come with an automatic caching logic, especially when the component relies on external services with a limited requests rate (e.g., public API accessed with free accounts with a limited amount of requests). For example, the localization component implements a caching mechanism to avoid repeating requests with the same labels, saving requests.

As a point of reference of how time-consuming can \tool be, we tested the tool on an Intel i7 laptop equipped with a Nvidia GeForce GTX 1050 and we were able to reconstruct 50 tweets in 20 seconds, adding demographic information with ~\newcite{wang2019demographic} and applying location disambiguation via the GeoNames Web Services.\footnote{\url{https://www.geonames.org/export/ws-overview.html}} Note that, however, some components are restricted by their own rate-limits (e.g., Twitter API v2) that might slow down the pipeline.

\begin{figure*}[ht!]
  \centering  \includegraphics[width=1\textwidth]{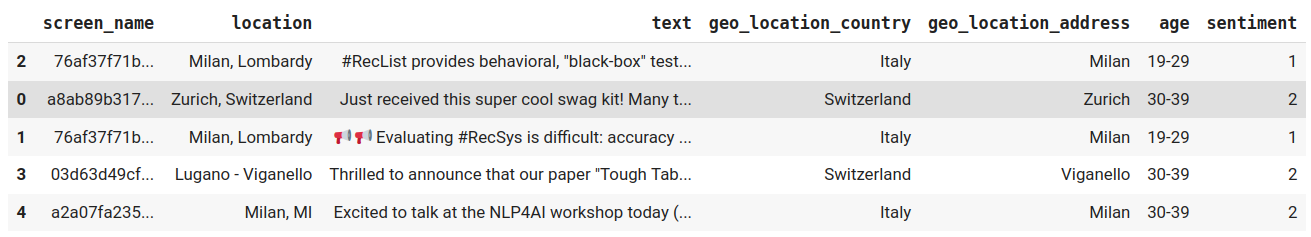}
  \caption{An example of a dataset enriched with sentiment analysis (2 is positive, 1 is neutral), location, age of the sender information. The `location' field, extracted with Twitter APIs, has been disambiguated and split into `geo\_location\_country' and `geo\_location\_address'. Screen names have been hashed (see Section \ref{sec:anonymity} for a discussion on privacy).}
  \label{fig:predicted}
\end{figure*}

Figure~\ref{fig:predicted} shows an example of a dataset enriched with sentiment prediction and location.

\subsection{Additional Features}

\tool exposes wrapping behaviors through the use of Python decorators to simplify the development of pipelines. For example, a common use case is to handle ``missing'' elements in the pipelines: a geolocalizer cannot be run if the user written location was not retrieved. This can break the pipeline (i.e., running the Geolocation on \texttt{None} generates an error). However, this is often not known at the start of the pipeline. This requires to write code to 1) temporarily skip data with missing text, 2) run the classifiers 3) return, to the caller, the entire dataset annotated with the new property where possible (to not compromise other steps). \tool exposes a simple decorator that automatically applies this kind of filtering, see Listing~\ref{listing:decorator}. The same functionality can be useful for a topic modeling pipeline or for a sentiment classifier.

\begin{listing}[h]
\begin{minted}[fontsize=\footnotesize]
{python}
@not_null("text")
def infer(self, data):

    [...]
    preds = model.predict(data["text"])

    return {"locations": preds}
        
\end{minted}
\caption{Extending class methods with decorators to support more complex behaviors. The `not\_null' decorator handles skipping null values so that the pipeline does not break during the flow.}
\label{listing:decorator}
\end{listing}

\subsection{Additional Resources}

\tool is available as Python package,\footnote{\url{https://github.com/MilaNLProc/twitter-demographer}} released under the research-friendly and open-source MIT license. It is also published on the PyPi repository,\footnote{\url{https://pypi.org/project/twitter-demographer/}} and can be installed with the \texttt{pip} package manager.
Automatic testing and deploying is handled via GitHub actions and the current state of the package can be checked online.\footnote{\url{https://github.com/MilaNLProc/twitter-demographer/actions}}
\tool also comes with online documentation that is available online at Read the Docs.\footnote{\url{https://twitter-demographer.readthedocs.io/en/latest/}} Tutorial notebooks are available on the GitHub repository. A video showcasing \tool usage can be found on YouTube.\footnote{\url{https://www.youtube.com/watch?v=NYljrfkLnU8}}

\subsection{Reproducibility}

The flow-based system supports reproducibility of data pipelines in a research environment. The pipeline itself, with the result, can be versioned into a JSON file for future machine-to-machine communication into data pipelines (inspired by~\newcite{chia2021beyond}). Moreover, component pipelines can be shared and used to augment the same or different datasets multiple times, reducing inconsistency that can arise when we reconstruct and enrich data.

\subsection{Privacy by design}
\label{sec:anonymity}
Inferring demographic attributes of users has many advantages for both data analysis and social science research, but it has obvious dual-use potential. I.e., ill-intentioned users could abuse it for their own gains. 
Users might have chosen not to disclose their information on purpose, so inferring them might go against their wishes. Given the ``right'' tools, we can also infer protected attributes. Moreover, collecting enough demographic attributes can identify real owners of individual users, or at least reduce the number of potential candidates substantially. The latter raises privacy concerns. 

Following the recommendations of the EU's General Data Protection Regulation \cite[GDPR,][]{GDPR}, we implement a variety of measures to ensure pseudo-anonymity by design. Using \tool provides several built-in measures to remove identifying information and protect user privacy:
1) removing identifiers,
2) unidirectional hashing, and
3) aggregate label swapping.

At the end of the reconstruction, we drop most of the personal information that we have been reconstructed (e.g., tweet id, profile URLs, images, and so on). Whenever possible, the information is anonymized. E.g.,  screen names are replaced with a globally consistent, but unidirectional hash code. In this way, we can retain the user-features mapping within the dataset (enabling further analysis, like aggregations), without allowing people to identify Twitter users (at least not without significant and targeted effort).
In addition, we randomly swap the complete set of labels of a subset of the final data, i.e., all labels attached to one instance are transferred to another instance, and vice versa. This procedure reduces the possibility of finding correlations between individual texts and their labels, which reduces its value for model training. However, we expect this use not to be a user priority. On the other hand, swapping does not affect aggregate statistics and the kind of analysis based on them.

\section{Conclusions}

We end this paper by discussing a broader outlook for this project and discussing the limitations that still affect \tool.

\subsection{Broader Outlook}
\label{sec:outlook}
We are constantly improving this library to support more use-cases and more models. For example, we are working on making the geolocation independent of third-party APIs like Geonames, trying to instead to support the download of Geonames index to query (thus improving speed and mitigating rate-limits). We are introducing multiple methods for topic modeling and additional components for text-clustering~\cite{grootendorst2020bertopic} and hyper-parameter optimization tools to find the optimal values for these.
We aim at providing a simple interface to address different user needs. While the tool is momentarily focused on Twitter, most of the components that we have been defined have a broader usage (e.g., the localization component).

\subsection{Limitations}

\tool does not come without limitations. Some of these are related to the precision of the components used; for example, the Geonames decoder can fail the disambiguation - even if it has been adopted by other researchers and services. At the same time, the the topic modeling pipeline can be affected by the number of tweets used to train the model and by other training issues (fixing random seeds can generate suboptimal solutions).

\tool wraps the API from~\newcite{wang2019demographic} for age and gender prediction. However, those predictions for gender are binary (male or female) and thus give a stereotyped representation of gender. Our intent is not to make normative claims about gender, as this is far from our beliefs. \tool allows using other, more flexible tools. The API needs both text and user profile pictures of a tweet to make inferences, for that reason \tool has to include such information in the dataset during the pipeline execution. While this information is public (e.g., user profile pictures), the final dataset contains also inferred information, which may not be publicly available (e.g., gender or age of the user). We cannot completely prevent misuse of this capability but have taken steps to substantially reduce the risk and promote privacy by design.

\section*{Ethical Considerations}
As outlined in Section \ref{sec:anonymity}, inferring user attributes carries the risk of privacy violations. We follow the definitions and recommendations of the European Union's General Data Protection Regulation for algorithmic pseudo-anonymity. We implement several measures to break a direct mapping between attributes and identifiable users without reducing the generalizability of aggregate findings on the data.
Our measures follow the GDPR definition of a ``motivated intruder'', i.e., it requires ``significant effort'' to undo our privacy protection measures. However, given enough determination and resources, a bad actor might still be able to circumvent or reverse-engineer these measures. This is true independent of \tool, though, as existing tools could be used more easily to achieve those goals.
Using \tool provides practitioners with a reasonable way to protect anonymity.


\bibliography{anthology,custom}
\bibliographystyle{acl_natbib}

\end{document}